\documentclass{svproc}

\usepackage{url}

\usepackage{graphicx}
\usepackage{booktabs}
\usepackage{multirow}
\usepackage{amsmath}
\usepackage{amssymb}
\usepackage{siunitx}
\usepackage{eurosym}
\usepackage{lmodern}
\usepackage[hidelinks]{hyperref}
\usepackage[backend=biber, style=numeric, sorting=none, backref=true]{biblatex}
\DeclareSourcemap{
  \maps[datatype=bibtex]{
    \map{
      \step[fieldset=abstract, null]
    }
  }
}

\begin{document}
\mainmatter              

\title{Learning to Modulate, Not to Cycle: Soft Actor--Critic Recovers
Inverter-Style Heat-Pump Control}
\titlerunning{~}

\author{Faizan Ahmed \and Aniket Dixit \and James Brusey}
\authorrunning{~}   
\tocauthor{Faizan Ahmed}

\institute{Centre for Computational Science and Mathematical Modelling,\\
Coventry University, United Kingdom\\
\email{ahmedf84@uni.coventry.ac.uk}}

\maketitle

\begin{abstract}
    On--off cycling is the main cause of compressor wear in residential heat pumps, yet reinforcement learning (RL) controllers for buildings typically optimise only energy cost and thermal comfort, ignoring how much the learned policy cycles. We add a levelised compressor-wear term to the control reward and study how the resulting behaviour depends on the RL algorithm. Training Soft Actor--Critic (SAC) and Proximal Policy Optimisation (PPO) on an identical Markov decision process for the BOPTEST \texttt{bestest\_hydronic\_heat\_pump} case, we find that SAC learns a continuous modulation policy that keeps the compressor permanently engaged---the operating principle of an inverter-driven heat pump---achieving zero start-ups per day, whereas PPO collapses to bang--bang control that cycles more than the baseline. On the BOPTEST emulator the SAC policy cuts thermal discomfort by up to \SI{90.7}{\percent} for an \SI{11.5}{\percent} cost increase, while eliminating all baseline cycling.
\end{abstract}

\section{Introduction}

Heat pumps are central to the decarbonisation of residential heating, but their delivered efficiency and service life depend strongly on \emph{how} they are operated, not only on \emph{how much} energy they consume. Single-stage units satisfy a heating load by switching the compressor fully on and off; every start incurs a transient during which refrigerant migration, oil return and pressure equalisation depress the coefficient of performance (COP), and the accumulated number of starts is a primary driver of mechanical wear and eventual failure~\cite{MADANI1,MADANI2,KARLSSON2007221}. The standardised part-load degradation coefficient $C_d$ used in rating procedures encodes precisely this cycling penalty~\cite{standard2008performance}. Inverter-driven (variable-capacity) heat pumps were introduced specifically to suppress cycling: by modulating compressor speed they track the load continuously, run for longer at lower capacity, and thereby raise seasonal efficiency while reducing the number of damaging start-ups~\cite{MADANI1,MADANI2,KARLSSON2007221}.

A controller for such equipment should therefore be judged on three axes simultaneously: thermal comfort, operating (electricity) cost, and \emph{control smoothness}, i.e.\ how much avoidable cycling it commands. Reinforcement learning is a popular model-free route to building HVAC control~\cite{WANG2020115036,VAZQUEZCANTELI20191072}, with supervised benchmarking through the open BOPTEST framework~\cite{Blum03092021}. Yet most RL control studies reward only a weighted sum of comfort violation and energy cost; the cycling behaviour of the learned policy is rarely measured and almost never optimised so that a policy can earn an excellent cost--comfort score while chattering the compressor in a way that real hardware could not sustain.

This paper closes that gap for hydronic heat-pump control. We make three contributions.

\begin{enumerate}
\item \emph{A smoothing-aware reward} that augments the comfort-plus-cost objective with a physically grounded, monetised compressor-wear term, levelising the compressor cost over its rated number of starts (Section~\ref{sec:reward})---making switching a first-class objective without an explicit minimum-runtime constraint.

\item \emph{An algorithm-class effect on smoothness:} holding the environment, reward and observations fixed, the RL algorithm determines the \emph{qualitative} form of the control signal. SAC learns a continuous-modulation policy that never switches off (zero starts/day, inverter-style operation), whereas PPO learns a bang--bang policy that cycles \emph{more} than the uncontrolled baseline.

\item \emph{A validated comfort--cost trade-off:} transferred to the BOPTEST emulator for this test case, the smooth SAC policy beats the built-in baseline on thermal discomfort by up to \SI{90.7}{\percent} for a modest cost increase, while removing all baseline cycling. \end{enumerate}

Our focus is the reward design, the resulting control behaviour, and its BOPTEST validation.

\section{Related Work}
\paragraph{RL for building and HVAC control:}
Data-driven building control has been surveyed extensively~\cite{WANG2020115036, VAZQUEZCANTELI20191072}, with deep RL built on actor--critic foundations for continuous control~\cite{pmlr-v80-haarnoja18b,schulman2017proximalpolicyoptimizationalgorithms}. Model-predictive control remains the strong supervised baseline~\cite{DRGONA2020190}, and hybrid RL--MPC schemes have also been explored~\cite{ARROYO2022118346}. BOPTEST~\cite{Blum03092021} standardises the comparison through a common emulator/KPI interface; we adopt its \texttt{bestest\_hydronic\_heat\_pump} case and KPI definitions.

\paragraph{Cycling and capacity control of heat pumps:}
The penalty of on--off operation is well established in the refrigeration literature: Madani et al.~\cite{MADANI1, MADANI2} and Karlsson and Fahl\'en~\cite{KARLSSON2007221} quantifies how variable-capacity (inverter) control reduces start-ups and raises seasonal performance, and rating standards encode the part-load cycling loss through the degradation coefficient~\cite{standard2008performance}. These results motivate treating the \emph{number of starts} as a control objective. Finally, that unconstrained RL produces high-frequency, oscillatory actions is a known failure mode addressed by explicit smoothness regularisers~\cite{MYSORE}; we show that for heat-pump cycling the \emph{algorithm class}---entropy-regularised off-policy learning---can itself supply the smoothness bias.

\section{Problem Formulation}
\label{sec:problem}
We formulate single-zone heat-pump control as a discrete-time Markov decision process (MDP) with a control step of $\Delta t = \SI{900}{\second}$ (15~min), matching the BOPTEST test case. Episodes span 14 days (\num{1344} steps).

\subsection{Action, Observation, and Heat-Pump Operating Principle}
The action is the scalar heat-pump command $u_{\mathrm{HP}}\in[0,1]$, the normalised compressor capacity, where $0$ is fully off and $1$ is rated capacity. The observation is a Case-D--aligned vector (zone temperature, ambient temperature, direct-normal irradiance, internal gains, the comfort set-points, the current price, and a sine/cosine time-of-day encoding) followed by an 8-step look-ahead forecast of the exogenous signals; two duty features (previous on/off state and normalised dwell time) are appended when the wear term is active.

The action semantics tie the learning problem to the hardware. A single-stage (on--off) unit admits only the endpoints $u_{\mathrm{HP}}\in\{0,1\}$ and meets partial loads by duty-cycling between them---each alternation a start. An inverter unit admits the whole interval and meets a partial load by sustaining an intermediate capacity, eliminating the alternation~\cite{KARLSSON2007221,MADANI1, MADANI2}. Our action space exposes the full interval, so the agent is \emph{free} to behave as an on--off thermostat (the endpoints) or as an inverter (the interior); the emerged behaviour is selected by the learning algorithm, and that selection is the object of study.

\subsection{Reward}
\label{sec:reward}
The per-step reward is the negative weighted sum of three integrated
costs over $\Delta t$:
\begin{equation}
\label{eq:reward}
r_t = -\big(\, w_{\mathrm{disc}}\, J^{\mathrm{disc}}_t
            + w_{\mathrm{cost}}\, J^{\mathrm{cost}}_t
            + w_{\mathrm{wear}}\, J^{\mathrm{wear}}_t \,\big).
\end{equation}

The thermal-discomfort term integrates the violation of the time-varying comfort band, $J^{\mathrm{disc}}_t = \big(\max(T^{\mathrm{lo}}_t - T^{\mathrm{zone}}_t, 0) + \max(T^{\mathrm{zone}}_t - T^{\mathrm{hi}}_t, 0)\big)\,\Delta t_h$, in kelvin-hours; the electricity-cost term is $J^{\mathrm{cost}}_t = P^{\mathrm{tot}}_t\,\Delta t_h\, \pi_t / 1000$ in euros, with $\pi_t$ the (highly dynamic) tariff. The novel compressor-wear term monetises switching. With a binary duty $d_t = \mathbb{1}[u_{\mathrm{HP},t} > \varepsilon]$ ($\varepsilon = 10^{-3}$), a \emph{start} is a rising edge $\max(d_t - d_{t-1}, 0)$, and each start is charged a levelised cost

\begin{equation}
\label{eq:wear}
J^{\mathrm{wear}}_t = c_{\mathrm{cyc}} \cdot \max(d_t - d_{t-1}, 0),
\qquad c_{\mathrm{cyc}} = \SI{0.0133}{\text{\euro}},
\end{equation}

obtained by amortising a representative compressor replacement (\SI{\sim2000}{\text{\euro}}) over its rated \num{\sim150000} starts. A short-cycle surcharge penalises starts separated by fewer than four steps. We deliberately express wear in the same currency (euros) as the energy term so that the weights $w$ have a transparent economic meaning; in all experiments below $w_{\mathrm{disc}} = w_{\mathrm{cost}} = w_{\mathrm{wear}} = 1$.

A key property of \eqref{eq:wear} is that a controller can drive $J^{\mathrm{wear}}$ to zero in two ways: by never running the compressor, or by running it \emph{continuously} so that no rising edge ever occurs. The first is excluded by the discomfort term (the zone would freeze); the second is exactly the inverter-modulation regime. Whether an RL agent discovers the second, benign optimum turns out to depend on the algorithm.

\section{Method}
\label{sec:method}
\paragraph{Training environment (surrogate plant):}
The agents are not trained on the BOPTEST emulator directly but on a fast, data-driven surrogate of the plant identified offline using Sparse Identification of Nonlinear Dynamics (SINDy) ~\cite{brunton2015discovering}. The surrogate advances the zone thermal state one control step at a time, mapping the current state, the heat-pump command $u_{\mathrm{HP}}$ and the exogenous inputs to the predicted next-step state; the agent's observation at each step is formed directly from this predicted state, so the full \num{5e6}-step training budget is collected at a small fraction of the cost of stepping the high-fidelity emulator. A companion map supplies the heat-pump electrical power entering the cost and wear terms of \eqref{eq:reward}. We adopt this surrogate as-is from our earlier  work~\cite{dixit2025learninglesssindysurrogates} and regard its identification as outside the present scope; the resulting policies are transferred to the genuine BOPTEST emulator for all evaluation reported below.

\paragraph{}
We train two model-free agents with Stable-Baselines3~\cite{JMLR:v22:20-1364}. \textbf{SAC}~\cite{pmlr-v80-haarnoja18b} is an off-policy, maximum-entropy actor--critic whose stochastic policy, entropy bonus and replay buffer reward exploration of the \emph{interior} of the action interval, biasing it away from the saturated rails. \textbf{PPO}~\cite{schulman2017proximalpolicyoptimizationalgorithms} is an on-policy clipped policy-gradient method for which the cheapest way to raise the average duty is to shift probability mass onto the $u_{\mathrm{HP}}=1$ and $0$ rails---a tendency towards bang--bang. Both agents share network sizes, discount ($\gamma=0.99$), learning rate ($3\times10^{-4}$), observation space and the reward of \eqref{eq:reward}; only the algorithm differs, and each is trained for \num{5e6} steps before out-of-sample evaluation on the genuine BOPTEST emulator.

\paragraph{Evaluation protocol:}
We evaluate on the two BOPTEST scenarios---the \emph{peak} and \emph{typical} heating periods, each 14 days under the highly dynamic tariff---against the test case's built-in baseline controller. We report (i) BOPTEST closed-loop KPIs; (ii) cycling metrics from the binary duty signal (switches and starts per day, and the wear of \eqref{eq:wear}); and (iii) statistics of the commanded signal. The baseline duty is inferred from its power trace ($P_{\mathrm{HeaPum}} > \SI{50}{\watt} \Rightarrow$ on).

\section{Results}
\label{sec:results}

\subsection{SAC Eliminates Cycling; PPO Worsens It}
Table~\ref{tab:cycling} reports the cycling behaviour. SAC records \emph{zero} switches and \emph{zero} starts on both scenarios, eliminating the \numrange{1.07}{1.50} daily starts of the baseline. PPO does the opposite, cycling \numrange{2}{4} times more often than the baseline and turning the wear term \emph{negative}. The same reward, optimised by two algorithms, yields opposite cycling outcomes.

\begin{table}[t]
\footnotesize
\centering
\caption{Compressor cycling and monetised wear over the 14-day scenarios.
Lower is better. ``Starts/day'' is the number of off$\to$on transitions;
wear follows Eq.~\eqref{eq:wear}; TV is the mean step-to-step change of
the command.}
\label{tab:cycling}
\setlength{\tabcolsep}{4pt}
\begin{tabular}{llrrrr}
\hline\noalign{\smallskip}
Scenario & Controller & Switch/d & Start/d & Wear (\euro/d) & TV \\
\noalign{\smallskip}\hline\noalign{\smallskip}
\multirow{3}{*}{Peak}
 & Baseline & 2.14 & 1.07 & 0.0143 & 0.021 \\
 & \textbf{SAC} & \textbf{0.00} & \textbf{0.00} & \textbf{0.0000} & 0.046 \\
 & PPO & 8.51 & 4.29 & 0.0570 & 0.096 \\
\noalign{\smallskip}\hline\noalign{\smallskip}
\multirow{3}{*}{Typical}
 & Baseline & 3.00 & 1.50 & 0.0200 & 0.018 \\
 & \textbf{SAC} & \textbf{0.00} & \textbf{0.00} & \textbf{0.0000} & 0.025 \\
 & PPO & 5.29 & 2.64 & 0.0352 & 0.055 \\
\noalign{\smallskip}\hline
\end{tabular}
\end{table}

\subsection{The Control Signal is Continuous, not Gated}
\begin{figure}[tbp]
\centering
\begin{minipage}[t]{0.5\textwidth}
\centering
\includegraphics[width=\textwidth]{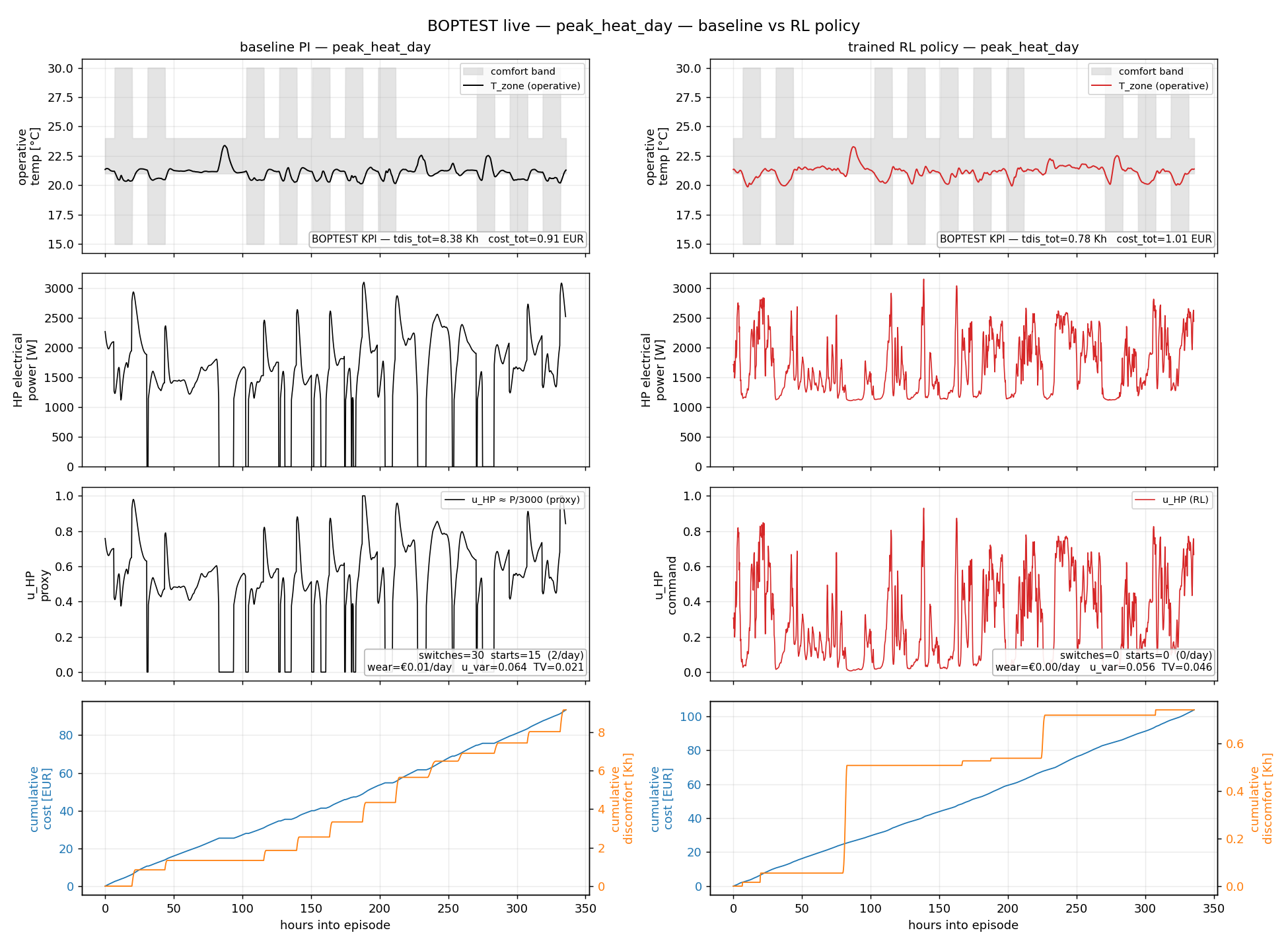}\\
{\small (a) SAC, peak day --- smooth modulation, 0 starts}
\end{minipage}%
\begin{minipage}[t]{0.5\textwidth}
\centering
\includegraphics[width=\textwidth]{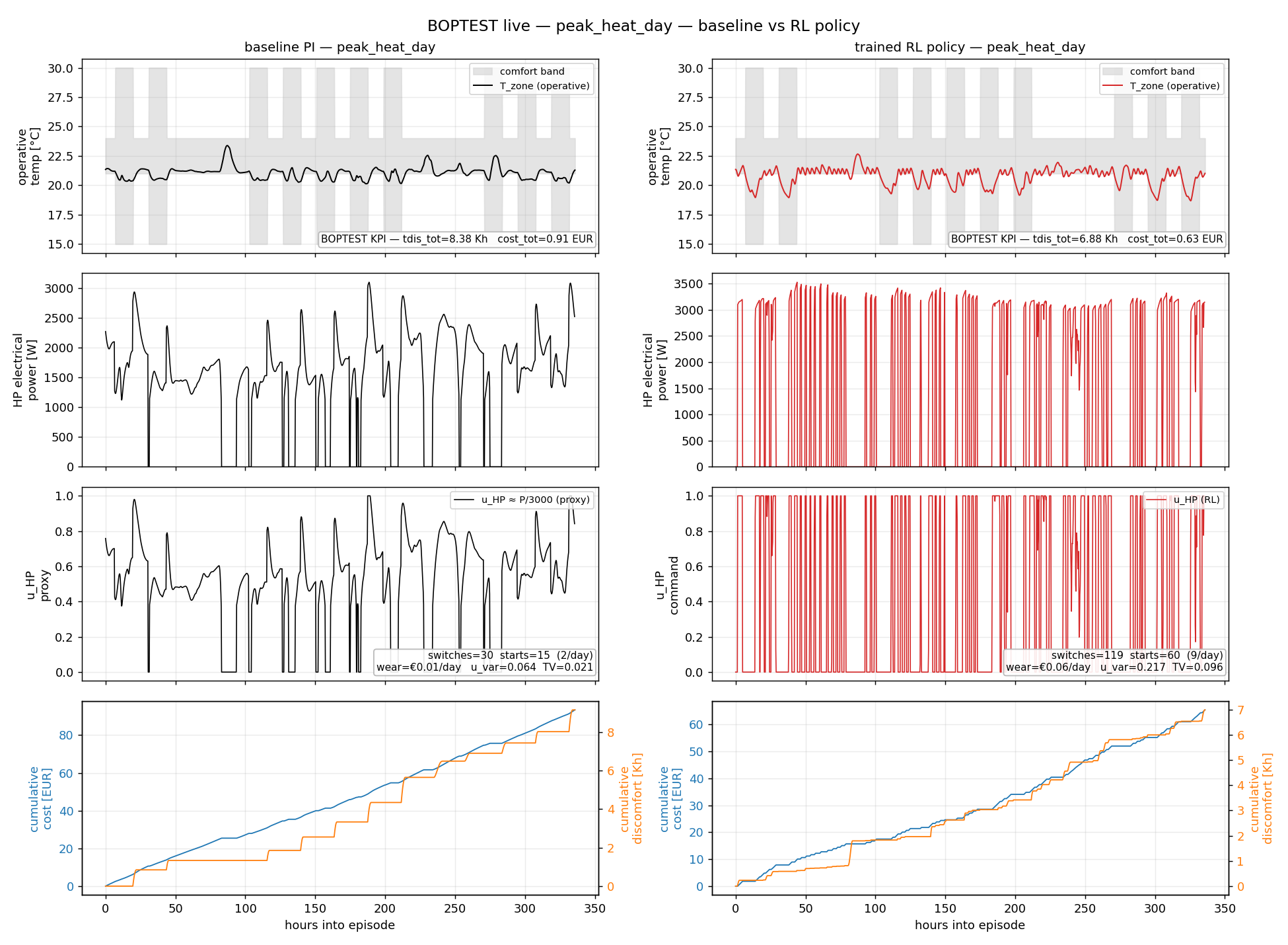}\\
{\small (b) PPO, peak day --- 4.29 starts/day}
\end{minipage}

\vspace{1.5mm}

\begin{minipage}[t]{0.5\textwidth}
\centering
\includegraphics[width=\textwidth]{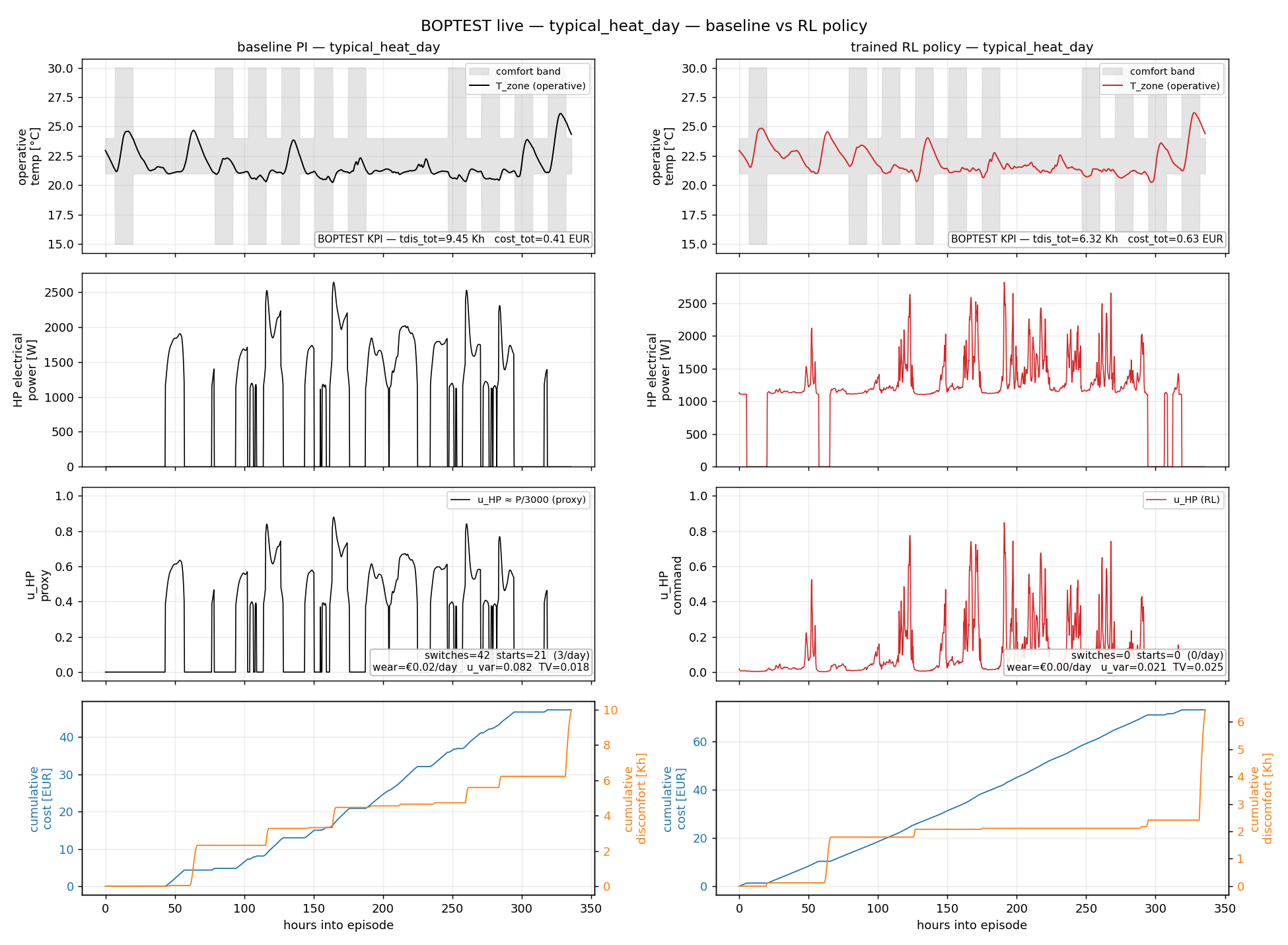}\\
{\small (c) SAC, typical day --- smooth modulation, 0 starts}
\end{minipage}%
\begin{minipage}[t]{0.5\textwidth}
\centering
\includegraphics[width=\textwidth]{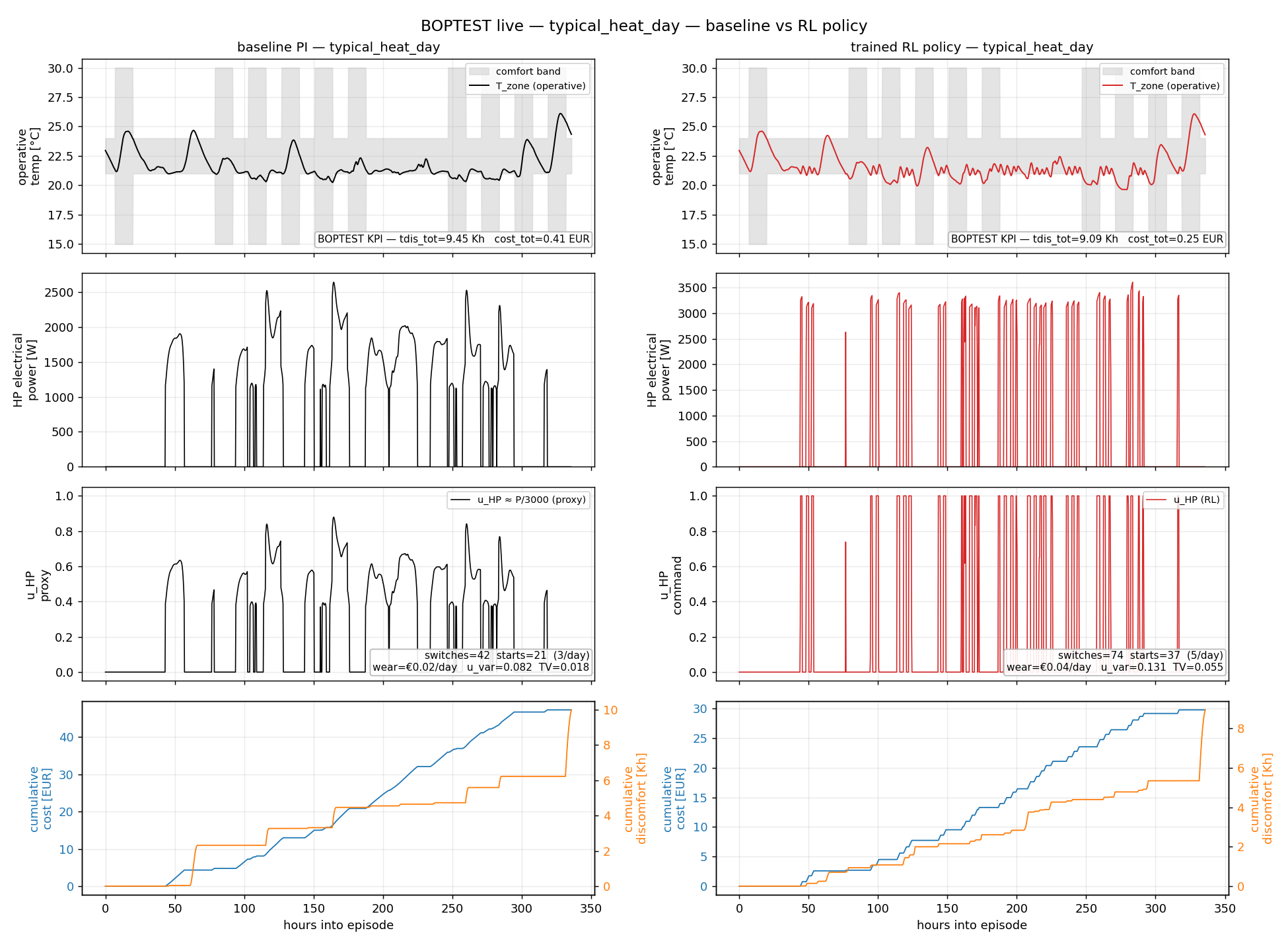}\\
{\small (d) PPO, typical day --- 2.64 starts/day}
\end{minipage}
\caption{Same reward, opposite controllers. Closed-loop traces on BOPTEST for the
peak (top) and typical (bottom) heating days; each panel shows zone
temperature against its comfort band (upper) and the commanded capacity
$u_{\mathrm{HP}}$ (lower). SAC (a,~c) modulates smoothly and never lets the
compressor restart---zero starts/day with comfort intact, learned inverter
operation even on the milder day. PPO (b,~d) slams $u_{\mathrm{HP}}$
between off and on, firing \num{4.29} and \num{2.64} starts/day, more than
the baseline.}
\label{fig:contrast}
\end{figure}

The commanded traces explain \emph{how} SAC reaches zero starts. SAC's $u_{\mathrm{HP}}$ varies substantially (non-zero TV in Table~\ref{tab:cycling}) yet \emph{never} drops to the off state: $u_{\mathrm{HP}}\in[0.007,0.93]$ with \SI{0.0}{\percent} of steps below the duty threshold on the peak day, and $[0.001,0.85]$ with again \SI{0.0}{\percent} off on the typical day---the compressor turns on once and then modulates continuously, the signature of inverter operation. PPO instead spends \SI{60.6}{\percent} (peak) and \SI{82.3}{\percent} (typical) of steps fully off and saturates at the $u_{\mathrm{HP}}=1$ rail---a bang--bang controller. This is the central empirical finding of the paper, and Fig.~\ref{fig:contrast} shows it directly for both the peak and typical days.

\subsection{Beating Discomfort on BOPTEST}
On the peak day (Table~\ref{tab:kpi}) SAC cuts thermal discomfort from \SI{8.38}{} to \SI{0.78}{\kelvin\hour} (a \SI{90.7}{\percent} reduction) for an \SI{11.5}{\percent} increase in operating cost, and reduces it by \SI{33.1}{\percent} on the typical day. The smooth policy thus \emph{purchases} comfort with a small, bounded cost premium while removing all cycling. PPO occupies the opposite corner: it lowers cost (it heats less) but barely improves comfort (\SI{17.9}{\percent} and \SI{3.8}{\percent}) and cycles heavily.

\begin{table}[t]
\footnotesize
\centering
\caption{BOPTEST closed-loop KPIs (14-day scenarios). $t_{\mathrm{dis}}$:
thermal discomfort (\si{\kelvin\hour}); cost (\euro); energy: source
energy (\si{\kilo\watt\hour}); emis: emissions (\si{\kilo\gram} CO$_2$).
Lower is better for discomfort, cost and emissions.}
\label{tab:kpi}
\setlength{\tabcolsep}{4pt}
\begin{tabular}{llrrrr}
\hline\noalign{\smallskip}
Scenario & Controller & $t_{\mathrm{dis}}$ & cost & energy & emis \\
\noalign{\smallskip}\hline\noalign{\smallskip}
\multirow{3}{*}{Peak}
 & Baseline & 8.382 & 0.909 & 3.478 & 0.581 \\
 & \textbf{SAC} & \textbf{0.777} & 1.013 & 3.860 & 0.645 \\
 & PPO & 6.884 & 0.631 & 2.436 & 0.407 \\
\noalign{\smallskip}\hline\noalign{\smallskip}
\multirow{3}{*}{Typical}
 & Baseline & 9.446 & 0.413 & 1.773 & 0.296 \\
 & \textbf{SAC} & \textbf{6.322} & 0.631 & 2.721 & 0.454 \\
 & PPO & 9.088 & 0.255 & 1.107 & 0.185 \\
\noalign{\smallskip}\hline
\end{tabular}
\end{table}

The energy column of Table~\ref{tab:kpi} locates SAC's cost premium: it draws \SI{11.0}{\percent} more source energy than the baseline, almost exactly its \SI{11.5}{\percent} cost increase, so the premium buys \emph{capacity}, not tariff mistiming---the modulating compressor simply delivers more heat. PPO's apparently low cost and emissions are instead an artefact of under-delivery: it leaves the zone cold and cycles to do so.

\section{Discussion}
\label{sec:discussion}

\paragraph{Why SAC recovers inverter operation?}
The wear term \eqref{eq:wear} has two zero-cost optima---never on, or always on---and the discomfort term rules out the first. Reaching the second requires holding $u_{\mathrm{HP}}$ strictly inside $(\varepsilon, 1]$ for the whole episode while tracking a time-varying load. SAC's maximum-entropy objective values a broad, stochastic action distribution and penalises collapse onto the deterministic rails, while its off-policy replay smooths the critic; interior, continuously-modulated actions thus become the path of least resistance, and the agent converges to the benign always-on optimum. PPO, lacking a comparable entropy floor and updating on-policy, finds it easier to raise the average duty by mixing fully-on and fully-off actions, landing in the bang--bang basin. SAC's policy is an inverter controller, learned rather than engineered~\cite{KARLSSON2007221,MADANI1, MADANI2}, while PPO's is a thermostat. That the zero-start result holds across two scenarios differing by an order of magnitude in mean capacity ($\overline{u}=0.287$ peak versus $0.098$ typical) indicates that continuous modulation is a robust attractor of the SAC objective, not an artefact of one operating point.

\paragraph{Trade-off and practical implications:}
The algorithms map onto opposite ends of the Pareto picture: SAC is comfort-seeking (a small, bounded cost premium for near-zero discomfort and zero wear) and PPO cost-seeking (under-heating to save energy); for a heat pump SAC's corner is the desirable one, since comfort is the primary obligation and eliminating cycling protects the most expensive component. Because the wear term is levelised in the same currency as operating cost, the two are directly comparable: over the 14-day peak scenario SAC avoids all \num{15} baseline starts, worth \SI{0.20}{\text{\euro}} of levelised wear at the assumed \SI{0.0133}{\text{\euro}} per start, which exceeds the \SI{0.10}{\text{\euro}} (\SI{11.5}{\percent}) operating premium---so the premium is recovered roughly twofold (break-even at \SI{0.007}{\text{\euro}} per start), and the \num{21} avoided starts on the typical scenario cover their premium likewise ($\approx\!1.3\times$, break-even \SI{0.010}{\text{\euro}} per start). Field controllers normally suppress cycling with hand-tuned timers and hysteresis; our result shows equivalent smoothness can arise endogenously by pricing each start, with $w_{\mathrm{wear}}$ as the single knob along the smoothness--cost frontier.

\paragraph{Limitations:}
The duty threshold $\varepsilon=10^{-3}$ is permissive: a controller need only hold a near-zero capacity to register as ``on,'' lowering the bar for the always-on optimum; a threshold at a real inverter's minimum stable modulation (\SIrange{10}{20}{\percent}) would be a stricter test. The per-start cost is small in absolute terms and serves as a \emph{lifetime} proxy rather than an operating expense. Finally, training uses an uncalibrated surrogate plant, so we rely on the out-of-sample BOPTEST KPIs (Table~\ref{tab:kpi}) for cost--comfort claims and on the measured duty signal (Table~\ref{tab:cycling}, Fig.~\ref{fig:contrast}) for smoothness claims.

\section{Conclusion}
We added a monetised compressor-wear term to a heat-pump control reward and showed that the RL algorithm then determines the qualitative form of the learned control law. SAC discovered a continuous-modulation policy that eliminates cycling, reproducing the operating principle of an inverter-driven heat pump, while PPO collapsed to a bang--bang policy that cycled more than the baseline; on BOPTEST the SAC policy cut thermal discomfort by up to \SI{90.7}{\percent} for a modest, bounded cost increase. For equipment whose lifetime is governed by switching, entropy-regularised off-policy learning is thus structurally aligned with the smoothness the hardware requires. Future work will sweep the wear weight to trace the switches-versus-(cost\,+\,discomfort) Pareto frontier and benchmark against an MPC baseline on BOPTEST.

%

\printbibliography

\end{document}